\documentclass[sigconf]{acmart}

\usepackage{stfloats,microtype,balance,enumitem}
\usepackage{color}
\usepackage{courier}
\usepackage{microtype}
\usepackage{chemarrow}
\usepackage{soul}
\usepackage{gensymb}
\usepackage{pifont}

\AtBeginDocument{%
  \providecommand\BibTeX{{%
    \normalfont B\kern-0.5em{\scshape i\kern-0.25em b}\kern-0.8em\TeX}}}

\copyrightyear{2024}
\acmYear{2024}
\setcopyright{licensedusgovmixed}\acmConference[HRI '24]{Proceedings of the 2024 ACM/IEEE International Conference on Human-Robot Interaction}{March 11--14, 2024}{Boulder, CO, USA}
\acmBooktitle{Proceedings of the 2024 ACM/IEEE International Conference on Human-Robot Interaction (HRI '24), March 11--14, 2024, Boulder, CO, USA}
\acmDOI{10.1145/3610977.3634974}
\acmISBN{979-8-4007-0322-5/24/03}

\acmSubmissionID{1477}

\begin{document}

\newcommand{\tool}{{\textit{Polaris}}}
\newcommand{\drawingboard}{{\textit{Drawing Board}}}
\newcommand{\planvisualizer}{{\textit{Plan Visualizer}}}
\newcommand{\Planvis}{{\textit{Plan-vis}}}
\newcommand{\Novis}{{\textit{No-vis}}}
\newcommand{\planvis}{{\textit{plan-vis}}}
\newcommand{\novis}{{\textit{no-vis}}}

\newcommand{\eg}[0]{\textit{e.g.,}}
\newcommand{\ie}[0]{\textit{i.e.,}}

\newcommand{\goalautomaton}{{$G_A$}}
\newcommand{\goaltree}{{$G_T$}}
\newcommand{\branchingplan}{{$\rho$}}
\newcommand{\tcmd}[2]{\textbf{{\texttt{#1}}}: \texttt{\textit{#2}}}
\newcommand{\predicate}[2]{[#1]}

\newcommand{\ltlG}{{\mathrm{\textsf{\textbf{G}}}}}
\newcommand{\ltlF}{{\mathrm{\textsf{\textbf{F}}}}}
\newcommand{\ltlU}{{\mathrm{\textsf{\textbf{U}}}}}
\newcommand{\ltlX}{{\mathrm{\textsf{\textbf{X}}}}}
\newcommand{\ltlS}{{\mathrm{\textsf{\textbf{S}}}}}
\newcommand{\ltlB}{{\mathrm{\textsf{\textbf{B}}}}}
\newcommand{\ctlE}{{\mathrm{\textsf{\textbf{E}}}}}

\newcommand\transition[1]{\xrightarrow{\tiny\raisebox{-0.5ex}[0pt][-0.5ex]{#1}}{}}

\newcommand{\todo}[1]{{\hl{#1}}}
\newcommand{\david}[1]{{\color{red}\textbf{David:} #1}}
\newcommand{\laura}[1]{{\color{blue}\textbf{Laura:} #1}}
\newcommand{\mak}[1]{{\color{green}\textbf{Mak:} #1}}

\renewenvironment{quotation}
{\list{}{\leftmargin=12pt
  \listparindent \parindent \itshape
  \itemindent \listparindent
  \rightmargin \leftmargin
  \parsep \parskip}%
  \item\relax\noindent\ignorespaces}
{\endlist}

\title{Goal-Oriented End-User Programming of Robots}


\author{David Porfirio}
\orcid{0000-0001-5383-3266}
\affiliation{%
  \institution{NRC Postdoctoral Research Associate}
  \institution{U.S. Naval Research Laboratory}
  \city{Washington}
  \state{DC}
  \country{United States}
  \postcode{20375}
}
\email{david.porfirio.ctr@nrl.navy.mil}

\author{Mark Roberts}
\orcid{0000-0003-2690-7658}
\affiliation{%
  \institution{U.S. Naval Research Laboratory}
  \city{Washington}
  \state{DC}
  \country{United States}
  \postcode{20375}
}
\email{mark.roberts@nrl.navy.mil}

\author{Laura M. Hiatt}
\orcid{0000-0001-5254-2846}
\affiliation{%
  \institution{U.S. Naval Research Laboratory}
  \city{Washington}
  \state{DC}
  \country{United States}
  \postcode{20375}
}
\email{laura.hiatt@nrl.navy.mil}

\renewcommand{\shortauthors}{David Porfirio, Mark Roberts, \& Laura M. Hiatt}

\begin{abstract}
\textit{End-user programming} (EUP) tools must balance user control with the robot's ability to plan and act autonomously.
Many existing task-oriented EUP tools enforce a specific level of control, \eg{} by requiring that users hand-craft detailed sequences of actions, rather than offering users the flexibility to choose the level of task detail they wish to express.
We thereby created a novel EUP system, \tool{}, that in contrast to most existing EUP tools, uses \textit{goal predicates} as the fundamental building block of programs. Users can thereby express high-level robot objectives or lower-level checkpoints at their choosing, while an off-the-shelf task planner fills in any remaining program detail.
To ensure that goal-specified programs adhere to user expectations of robot behavior, \tool{} is equipped with a \planvisualizer{} that exposes the planner's output to the user before runtime.
In what follows, we describe our design of \tool{} and its evaluation with 32 human participants.
Our results support the \planvisualizer{}'s ability to help users craft higher-quality programs. Furthermore, there are strong associations between user perception of the robot and \planvisualizer{} usage, and evidence that robot familiarity has a key role in shaping user experience.
\end{abstract}





\maketitle
\section{Introduction}\label{sec:introduction}

\begin{figure}[!t]
    \centering
    \includegraphics[width=\columnwidth]{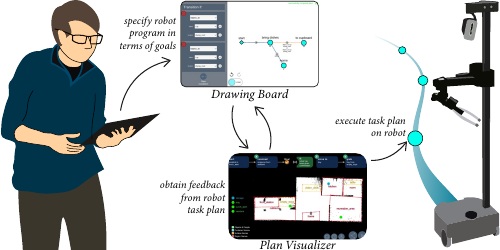}
    \caption{With \tool{}, end-user programmers  specify \textit{goal automata} and view the resulting plan in the \planvisualizer{}.}
    \label{fig:teaser}
\end{figure}

As robots permeate our daily lives, there is a growing demand for efficient and reliable approaches that allow end users to specify tasks for these robots to perform.
\emph{End-user programming} (EUP) tools, \ie{} software environments that enable these users to create and customize robot applications, represent a viable class of solutions.
Given the autonomous capabilities of everyday robots, users should be free to omit certain details from their programs. To illustrate, consider the following scenario: \textit{a caregiver needs a robot to deliver lunch to a resident in a care facility.} Rather than requiring the caregiver to specify a long string of actions (\eg{} \textit{go to cafeteria}, \textit{pick up tray}, \textit{go to food station}, \textit{wait for food}, \textit{go to resident}, and \textit{give food}), the caregiver can leverage the robot's ability to plan and act autonomously if simply given a desired goal state: \textit{lunch delivered}.

At the same time, these users must have the flexibility to express additional detail as needed,  based on their own domain expertise.
We define \textit{flexibility} as being able to choose between \textit{low} oversight (expressing minimal goals and letting the robot resolve the details) and \textit{high} oversight (specifying more details to constrain the robot).
Caregivers, in particular, can benefit from being able to access different levels of oversight \cite{stegner2022designing}.
Perhaps the caregiver in our example desires higher oversight due to their domain knowledge---the resident is usually in the recreation area midday, but they must be in their room to eat lunch due to care facility rules. In this case, there is an additional implied outcome that the resident should be in their room before the food is delivered. Constraining the robot with two goals in sequence will suffice: (1) \textit{resident alerted} to ensure that the resident knows to travel to their room, and then (2) \textit{lunch delivered}.

Unfortunately, there has been limited exploration of EUP tools that leverage robot autonomy while still affording users flexibility in program specification.
Most existing EUP tools necessitate high oversight by requiring users to hard-code robot actions, which as evidenced by existing datasets of user-generated action sequences, exhibits high contextual conformity \cite{liao2019synthesizing}.
In this work, we challenge the action-oriented EUP paradigm for human-robot interaction (HRI) by proposing \textit{goal predicates  } as an alternative fundamental building block of robot programs.
By selecting and parameterizing goal predicates, users can omit details on how an \textit{intended effect} (\ie{} ``goal state,'' or ``goal'' for brevity) is achieved. 
Furthermore, in recognizing that goals do not contain explicit information about which actions the robot will perform, we ask how goal-oriented EUP tools can ensure that user expectations match robot performance.

To address these gaps, we created a goal-oriented EUP system, \tool{}, that represents our vision of flexibility, abstracting away unnecessary detail while still affording users appropriate control over the robot.
Figure \ref{fig:teaser} depicts the high-level usage flow of \tool{}, which exists as a handheld tablet interface.
With \tool{}, end users specify \textit{goal automata}---a flow-based representation in which nodes in the flow represent goals rather than actions.
This enables end-user programmers to specify programs at a level of detail with which they are comfortable or that is required by their domain expertise.
\tool{} then automatically generates a branching task plan through off-the-shelf AI planning approaches.
To ensure that plans match developer intent and to provide feedback for refinement, \tool{} includes a \planvisualizer{} interface that exposes the plan to users.

The \tool{} system represents an ongoing research effort.
This paper describes a snapshot of this effort, culminating in \tool{} V1.0, and highlights our motivations and initial design decisions.
Our evaluation tests these design decisions, finds evidence that the \planvisualizer{} improves plan quality, and uncovers associations between user experience and both \planvisualizer{} usage and self-reported robot familiarity.
We conclude by offering design implications and discussing how these implications inform our own future work and future development of EUP systems in general.

Our contributions include:
\emph{Systems} --- the \tool{} system, a novel goal-oriented EUP tool and our primary contribution.
\emph{Empirical} --- an evaluation of \tool{}, namely the \planvisualizer{}'s ability to assist end-user programmers with creating goal-oriented programs.
\emph{Design} --- design implications that emerged from our evaluation.

\section{Related Work}
\tool{}' contributions and novelty are situated within end-user programming and draw heavily from goal-oriented task specification and automated planning in HRI.

\subsection{Robot End-User Programming}\label{subsec:related_work_eud}

\begin{table}
\small
\centering
\begin{tabular}{r|ccccc}
 & Front-End & AI & & Interface & User \\
 & Goal Preds & Planning & Domain & Feedback & Study \\
 \hline
 
 \textit{\textbf{Polaris}} & \checkmark & \checkmark & Service & \checkmark & \checkmark \\

 \textit{``Spbd''} \cite{brageul2008} & \checkmark &  & Nav. & \checkmark & \\
 
 \textit{JESSIE} \cite{kubota2020jessie} & & & Social & & \checkmark \\

 \textit{Tabula} \cite{porfirio2023sketching} & & \checkmark & Service & \checkmark & \\

 \textit{RoVer} \cite{porfirio2018authoring} & & & Social & \checkmark & \checkmark \\

 \textit{RoboFlow} \cite{alexandrova2015roboflow} & & & Service & & \checkmark \\

 \hline
\end{tabular}
\caption{A comparison of \tool{} to closely related EUP tools. Polaris exposes goal predicates to users, explicitly incorporates AI planning techniques, provides visual feedback on user programs through its interface, and is user-evaluated.}
\label{table:relatedwork}
\end{table}

\textit{End-user programming} pertains to the creation of software applications by the application users themselves \cite{barricelli2019end}.
Contributions in robot EUP often focus on novel ways to capture user intent through visual programming environments \citep[\eg{}][]{leonardi2019trigger, schoen2022coframe}, augmented reality \citep[\eg{}][]{cao2019vra, cao2019ghostar}, natural language \citep[\eg{}][]{forbes2015robot, gorostiza2011end}, or multimodal input \citep[\eg{}][]{porfirio2021figaro, beschi2019capirci}, to name a few examples.
In HRI, end-user programmers\footnote{We often refer to end-user programmers as simply ``end users'' or ``users'' for brevity.} are typically (though not always) programming novices, and may also be domain experts specialized in specific fields \cite{ajaykumar2021survey}.
\tool{}' target end-user programmer includes domain experts in need of personalized (\eg{} through goal-oriented specification) yet reliable (\eg{} through the \planvisualizer{}) robot execution, such as caregivers, military personnel, and disaster response teams.

\tool{}' contribution lies primarily in its \textit{goal-oriented} programming paradigm---users specify an intended effect in terms of goal state rather than an \textit{action-oriented} description of robot behaviors to achieve that effect.
Overwhelmingly, existing EUP systems for HRI are \textit{action-oriented}.
Action-oriented examples from the EUP literature include \textit{block-based} \citep[\eg{}][]{huang2016design, chung2016iterative, huang2017code3, schoen2022coframe}, \textit{flow-based} \citep[\eg{}][]{pot2009choregraphe, alexandrova2015roboflow}, and \textit{event-based} \citep[\eg{}][]{leonardi2019trigger} tools, in which the fundamental building block of a robot application is an action or command. 

Table \ref{table:relatedwork} characterizes \tool{}' novelty against a representative selection of similar, existing EUP systems and programming approaches.
Notably, \citet{brageul2008}'s \textit{simple programming by demonstration} (``\textit{spbd}'') interface is similar to \tool{}' goal-oriented nature in that it allows users to directly manipulate goal predicates. Unlike \tool{}, however, \textit{spbd} is limited to navigation domains and lacks a user study. Another tool, \textit{JESSIE}, similarly captures goal state within its program logic, but this logic is not exposed to the user \cite{kubota2020jessie}.
\tool{} additionally distinguishes itself from prior work that views goals as high-level task commands (\eg ``open a sliding door'' as a goal in \cite{aguinaldo2022robocat}) due to our strict definition of goals as expressing desired state rather than any information about the robot's actions.  

Goal-oriented nature aside, \tool{} draws heavily from other prior work.
\textit{Tabula}, in particular, exists within a handheld tablet, invokes a planner to determine robot behavior, and offers plan feedback through its user interface \cite{porfirio2023sketching}. 
\textit{Tabula}, however, only exposes actions to users and is not yet evaluated in a user study.
Although not incorporating a planner, both \textit{RoVer} \cite{porfirio2018authoring} and \textit{RoboFlow} \cite{alexandrova2015roboflow} afford users a similar flow-based specification interface to \tool{} and check pre and postconditions between consecutive robot actions. 
\textit{RoVer} is additionally similar to \tool{} in its user interface, namely through the inclusion of a dedicated feedback pane.

\subsection{Goal-Oriented Specification Paradigm}

Goals are a critical component of many formal representations, architectures, and models for autonomous agents.
The \textit{belief-desire-intention} (BDI) paradigm presents one such modeling approach for agent reasoning \cite{bratman1987intention} and has led to numerous agent-oriented programming languages, including \textit{AgentSpeak} \cite{rao2005agentspeak} and variants of \textit{CAN} \cite{sardina2011bdi}, of which goals are of great importance.
Goals are additionally critical to the specification of both classical and hierarchical planning problems \cite{ghallab2016automated, shivashankar2012hierarchical}.
For expressing robot programs purely via goals, \tool{} utilizes an approach most similar to \textit{Agent Planning Programs} \cite{DEGIACOMO201664}, in which programs are represented as transition systems with transitions between program states labeled by goals and guard functions.  

Prior work demonstrates how goal-oriented languages improve user outcomes.
In particular, \citet{hu2012teaching} shows how decomposing a task into higher-level objectives (``goals'') combined with block-based programming can improve learning outcomes among students.
Additionally, \citet{cox2007mixed} contrast two mixed-initiative planning approaches, \textit{goal manipulation} versus \textit{search} (\ie{} searching through action pre and postconditions), and find that goal manipulation surpasses search in terms of user performance and efficiency. These results support our design choice to expose goals to users through \tool{}. 

Although less common in robot EUP, goal-oriented specification is popular in related fields.
Prior work within the Internet of Things (IoT), in particular, has produced numerous EUP tools, languages, and architectures for specifying smart home configurations in terms of predicates (\eg{} \textit{temperature is} $x$, in which $x$ is a value in degrees) \cite{mayer2016smart, kovatsch2015iot}.
An underlying motivation of goal-oriented specification in IoT is to reduce development time \cite{noura2018growth}.

\subsection{Automated Planning in HRI}
\textit{Planning and acting} involves selecting \textit{what} an agent does and \textit{how} it does it \cite{ghallab2016automated}.
The scope of our work is on the former---\textit{what}.
Within this scope, \textit{automated} planning solves for plans with respect to a \textit{planning domain}, that is, constraints which are often specified in languages such as the \textit{Planning Domain Definition Language} (PDDL) \cite{fox2003pddl2} or the \textit{Action Notation Modeling Language} (ANML) \cite{smith2008anml}.  
Notable planning work in HRI involves reasoning about and responding to human behavior \cite{petrick2013planning, alili2009task,9812236} and creating robot planning domains through demonstration \cite{liang2019end}.
Recently, \citet{chakraborti2019plan} investigated the use of plan explanations to improve the shared understanding of a robot's decisions.

Various interfaces exist for visualizing and enabling end users to interact with plans, \textit{PDSim} being a notable example in robotics \cite{de2021automated}.
Of the existing planning interfaces in HRI, \textit{Tabula}  is most similar to \tool{} but is action-oriented and focuses more on the mechanism for capturing the intent of end-user programmers \cite{porfirio2023sketching}. A plethora of other such interfaces (\eg{} \texttt{RADAR-X} \cite{valmeekam2022radar}) exist outside of HRI.

\section{System Design}\label{sec:proto}

\begin{figure*}[!t]
    \centering
    \includegraphics[width=\textwidth]{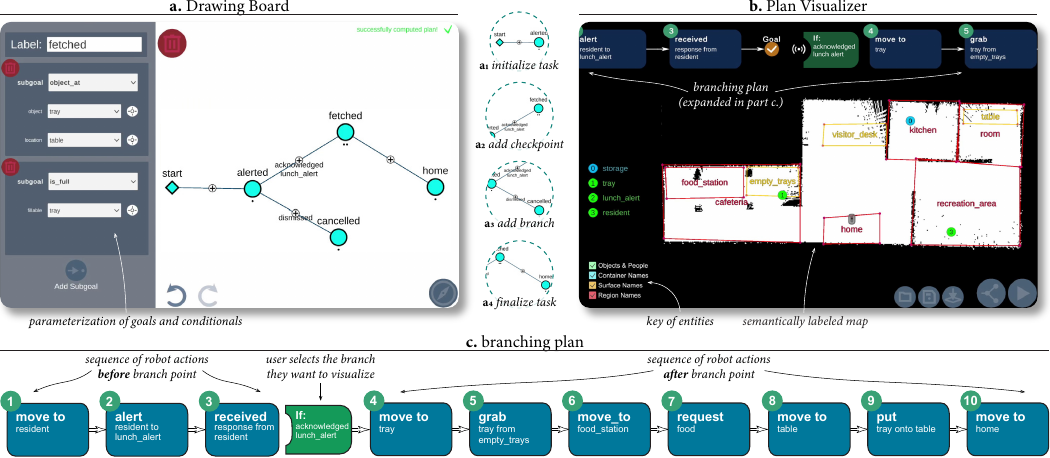}
    \caption{The \tool{} user interface, which includes (a) the \drawingboard{} for specifying goal automata and (b) the \planvisualizer{} for displaying the robot's plan. (c) A single branch of the branching plan from the caregiving scenario is displayed.}
    \label{fig:ui}
\end{figure*}

Our description of \tool{} begins by elaborating on the caregiving scenario presented in \textit{Introduction} (\S\ref{sec:introduction}) to illustrate the user's perspective, followed by our technical approach for
(1) specifying task objectives in terms of goals, (2) generating a task plan, (3) viewing the plan, (4) running the plan, and (5) \tool{}' implementation. 

\subsection{User Perspective}\label{sec:example}
Figure \ref{fig:ui} depicts the user's perspective of \tool{} with its various components described below.

\noindent{\textbf{World. }}The user begins by requesting a two-dimensional map of the environment from the robot and uploading semantic labels for key entities that the robot can recognize.
The semantically labeled map (accessible though the \planvisualizer{}, Figure \ref{fig:ui}b) depicts the \textit{world} that the robot operates within, and includes the two-dimensional representation of the robot's environment and the entities therein. 
There are five general categories of entities---objects, containers, surfaces, regions, and people.
Objects include anything that the robot can grab.
Containers include anything within which an object can be placed.
Surfaces are areas upon which objects can be placed and are non-traversable by the robot.
Regions are traversable areas in the environment.
People include the robot's potential interaction partners.
Within the world, the locations of objects and people represent initial positions, and these entities can be moved around throughout the course of a program's execution.

Figure \ref{fig:ui}b depicts the world within our caregiving scenario.
The care facility has been labeled with regions such as the \textit{recreation area}, the resident's \textit{room}, and the \textit{cafeteria}.
The robot can place items onto and remove items from surfaces such as the \textit{empty trays} surface and the resident's \textit{table}.
The \textit{tray} is a manipulable object and the \textit{resident} is an interactable person in the environment.

\noindent{\textbf{Creating a Goal Automaton. }} The user enters the \drawingboard{} to specify a program in terms of goals.
Figure \ref{fig:ui}a depicts the \drawingboard{} with an example user-based solution, called a \textit{goal automaton}.
In the goal automaton, blue nodes represent \textit{checkpoints}.
Checkpoints contain goals, thereby indicating the state of the world that the user wants the robot to achieve at that point in the program.
Connecting checkpoints with lines (called \textit{transitions}) enforces an ordering and allows users to specify a \textit{conditional}, or world state that must be true (possibly outside of the robot's control) for the robot to proceed from one checkpoint to another.
Checkpoints contain no indication of the robot's geographical location and are purely intended to represent program flow.

Figure \ref{fig:ui}a$_{1-4}$ shows the process of building the goal automaton to represent the caregiver's objectives.
The resident must be informed of lunchtime before lunch is served, so the caregiver's first step (Figure \ref{fig:ui}$a_1$) is to draw a new checkpoint and label it \textit{alerted}.
When a new checkpoint is created, a parameterization menu appears that prompts users to add goals to the checkpoint.
Users may also click on existing checkpoints during the course of goal automata creation to modify these checkpoints' goals.
Within \textit{alerted}, the caregiver assigns $x$ and $y$ values to the ``\textit{x}\textbf{\textit{--alertedTo--}}\textit{y}'' predicate to create a \textit{ground predicate} and assert a goal: The resident has been alerted to lunch being served, namely (\textit{resident}\textbf{\textit{--alertedTo--}}\textit{lunchtime}).\footnote{Our goal notation parenthesizes \textit{\textbf{bolded}} predicate symbols and \textit{italicized} terms.}

Next (Figure \ref{fig:ui}$a_2$), the caregiver draws a line from \textit{alerted} to a new checkpoint, \textit{fetched}, and inserts goals (\textit{tray}\textbf{\textit{--at--}}\textit{table}) and (\textit{tray}\textbf{\textit{--is full}}), indicating that after alerting the resident, lunch must be delivered.
The caregiver labels the transition from \textit{alerted} to \textit{fetched} with the ground predicate (\textbf{\textit{acknowledged--}}\textit{lunchalert}), in this case indicating a conditional that the robot can only proceed from the \textit{alerted} checkpoint to the \textit{fetched} checkpoint if the resident acknowledges the alert.
Throughout the course of goal automata creation, the user may click on existing transitions to modify the transitions' conditionals.
In order to handle the edge case in which the robot's lunchtime alert is dismissed (\eg{} if another caregiver has already served the resident lunch and wishes to cancel the robot's task), the next step taken by the caregiver (Figure \ref{fig:ui}$a_3$) is to draw a transition from \textit{alerted} to a new checkpoint, \textit{cancelled}, assign a goal of (\textit{\textbf{robotAt--}}\textit{home}), and label the new transition with the conditional \textit{\textbf{dismissed}}.
Finally (Figure \ref{fig:ui}$a_4$), the caregiver draws a new checkpoint from \textit{fetched} called \textit{home} and adds another (\textit{\textbf{robotAt--}}\textit{home}) goal.
More information on the semantics of goal automata can be found in \textit{Representing Goal Automata} (\S\ref{sec:drawingboard}).

\noindent{\textbf{Viewing and Running a Plan. }}As the user makes progress on their goal automaton, \tool{} computes a task plan behind the scenes, which contains the exact actions that the robot plans to take to achieve each of the caregiver's goals in sequence.
\tool{} presents this information to users via the \planvisualizer{} interface (Figure \ref{fig:ui}b).
Figure \ref{fig:ui}c depicts the plan based on the caregiver's goal automaton as it is presented to the user.
At any point during goal automaton creation, the user can flip back and forth between the \drawingboard{} and \planvisualizer{} to iterate on receiving plan feedback and performing modifications to their goal automaton.
Once the user presses the play button (Figure \ref{fig:ui}b, bottom right), the robot begins to execute the plan.
Branching plan creation, viewing, and execution are detailed further in \textit{Creating a Branching Plan} (\S\ref{sec:backend}), \textit{Viewing the Branching Plan} (\S\ref{sec:planvis}), and \textit{Plan Execution} (\S\ref{sec:exec}).

\subsection{Representing Goal Automata}\label{sec:drawingboard}

Formally, a goal automaton is a transition system \cite{baier2008principles} that guides the robot in achieving goals during its task and is represented by the tuple ($P$, $C$, $L_c$, $\longrightarrow$, $c_0$):

\setitemize{leftmargin=1.0em}
\begin{itemize}
  \setlength\itemsep{0.4em}
  \item [] \textbf{Predicates. }$P$ is a set of \textit{ground predicates}, \ie{} predicates with assigned variable values. Predicates primarily represent goals, but can also represent conditionals, namely world state that must be true for the robot to proceed. Intuitively, conditionals indicate that the robot must wait for a particular outcome that may be out of the robot's control.
  \item [] \textbf{Checkpoints. }$C$ is a set of \textit{checkpoints}. Intuitively, checkpoints represent points in the program in which the robot has achieved a desired set of goals $p \in 2^P$, in which $2^P$ is the power set of $P$.
  \item [] \textbf{Goals. }$L_c: C \rightarrow 2^{P}$ maps checkpoints to goals.
  \item [] \textbf{Transitions. }$\longrightarrow \subseteq C\times 2^{P} \times C$ is the transition relation between checkpoints subject to a conditional being true. For example, $p\in 2^P$ is a conditional within the transition $c_i \transition{$p$} c_j$. Intuitively, a transition labeled with conditional $p$ means, ``wait for $p$ to be true before transitioning between $c_i$ and $c_j$.'' A transition with no conditional annotation (\eg{} $c_i \transition{} c_j$) means ``transition from $c_i$ to $c_j$ if no other transitions from $c_i$ are able to be taken.''
  \item [] \textbf{Initial Checkpoint. } $c_0$ is the always-empty ``start'' checkpoint. $c_0$ represents initial state and does not have goals: $L_c(c_0) = \emptyset$.
\end{itemize}

Figure \ref{fig:ui}a depicts the interface for specifying goal automata, the \drawingboard{}.
Initially, the \drawingboard{} contains the empty checkpoint $c_0$.
Each new checkpoint $c_j$ must be drawn as $c_i \xrightarrow[]{} c_j$ such that $c_i \in C$ (\ie{} new checkpoints must connect to existing checkpoints), $L_c(c_j)=\emptyset$ (\ie{} new checkpoints \textit{initially} contain no goals), and new transitions \textit{initially} contain no conditionals.
While the intention is to support loops in future versions of \tool{}, goal automata are presently drawn as trees.

\subsection{Creating a Branching Plan}\label{sec:backend}

Branching plans are \textit{compiled} in real-time as changes are made to the goal automaton, provided that there are no underspecified transitions (\ie{} two transitions with the same conditional extending from the same checkpoint). If a checkpoint in the goal automaton contains conflicting goals (\eg{} the user asserts that a single-arm robot must hold two items at the same time), \tool{} will omit that checkpoint and further checkpoints in its subtree from compilation.

Formally, a branching plan is similar in tree structure to the goal automaton but consists of \textit{actions} rather than goals.
Let $A$ be the set of tree nodes in the plan and $L_a: A \rightarrow 2^P$ be the world state after a node's action has been executed.
Let $\longrightarrow \subseteq A\times 2^{P} \times A$ be the transition relation between nodes.
Given a goal automaton, \tool{} creates a branching plan such that there is an injective non-surjective mapping between checkpoints and plan nodes $f_m: C \rightarrow A$. For convenience, let $f_s: A \rightarrow 2^A$ map node $a_i$ to the set of nodes in $a_i$'s subtree. Subject to the following additional constraints, \tool{} leverages an off-the-shelf planner for plan creation.

\begin{figure}[!t]
    \centering
    \includegraphics[width=\columnwidth]{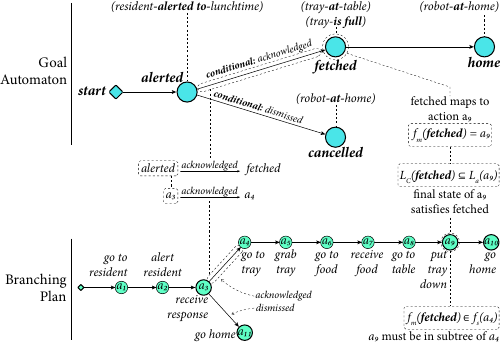}
    \caption{Computing branching plans from goal automata.}
    \label{fig:plan_creation}
\end{figure}

\bigskip
\noindent\textit{Constraint on Transitions}---For each transition $c_i \transition{$p$} c_j$ in the tree-like goal automaton, there must be a corresponding transition in the branching plan $a_i \transition{$p$} a_j$ such that $f_m(c_i) = a_i$ and $f_m(c_j) \in f_s(a_j)$, that is, $c_i$ maps to  $a_i$ and the mapping of $c_j$ is in the subtree of $a_j$.
Figure \ref{fig:plan_creation} illustrates this constraint---the transition between \textit{alerted} and \textit{fetched} maps to the transition between $a_3$ and $a_4$, and \textit{fetched} maps to a descendent of $a_4$.

\bigskip
\noindent\textit{Constraint on Goal Achievement}---A checkpoint's goals must match the state of the world after the completion of its corresponding action in the branching plan: $L_c(c) \subseteq L_a(f_m(c))$. Figure \ref{fig:plan_creation} illustrates this constraint: The goals of \textit{fetched} match the end effect of its corresponding action in the plan, $a_9$.

\subsection{Viewing the Branching Plan}\label{sec:planvis}

At any point during the creation of a goal automaton, end users may view the branching plan computed by \tool{} within the \planvisualizer{} interface.
The \planvisualizer{} draws from existing ``timeline'' interfaces in HRI \cite{sauppe2014design, authr2020schoen} in that it displays one branch of the plan at a time from left to right,  and within a horizontal scrollable pane overlaying the semantically labeled map.
Initially, the plan is displayed up to when a conditional is encountered. Users then select the conditional corresponding to the branch they wish to visualize via a dropdown menu. Following the user's selection, the \planvisualizer{} displays the corresponding branch up to the next conditional. Actions within each branch can be clicked to depict the world state that results from the execution of the clicked action.

\subsection{Plan Execution}\label{sec:exec}
When the user is satisfied with their goal automaton and resulting plan, they may execute the plan on the robot.
During plan execution, \tool{} enters a feedback-execution loop with the robot.
Rather than sending the robot actions directly from the plan, \tool{} converts actions to goals (\ie{} by using the end effects of an action).
This enables the robot to compute a new plan to achieve the end effect of each action, rendering the robot flexible to minor perturbations in the environment.
The robot is thereby able to re-plan when its perceived state of the world changes and repeat this process until the goal-converted action has been achieved.
The robot then sends a confirmation back to \tool{}, which converts the next action in the plan to a goal and sends the new goal to the robot.

\subsection{Implementation}\label{sec:implementation}
\tool{} exists within a front-end tablet and a back-end planner communicating over a RESTful API. 
The front end is implemented in Unity version 2022.2.1f1 \cite{unity} and is compiled to Android.
While \tool{} is primarily intended for handheld use, its implementation in Unity has enabled us to deploy it on a web browser and as a desktop application.

The \tool{} back end is implemented within Python 3.8 and accesses a planning domain expressed in PDDL \cite{fox2003pddl2}.\footnote{Planning domains are interchangeable within \tool{}. Example planning domains can be found at \url{https://osf.io/ewfd5/}.}
At a high level, the planning domain consists of (1) a set of predicates that operate over both the robot's state and entities in the world, and (2) a set of operations that the robot can perform, including how these operations affect both the robot and the world. We integrated an off-the-shelf planner, \textit{Fast Downward} \cite{helmert2006fast, helmert2006fasttranslator}, within the back end. 

We use the Hello Robot Stretch RE2 robot \cite{kemp2022design} as our runtime platform.
Communication between \tool{} and the robot occurs through the Noetic version of the Robot Operating System \cite{quigley2009ros}.
\section{System Evaluation}\label{sec:evaluation}

\begin{figure*}[t!]
    \includegraphics[width=\textwidth]{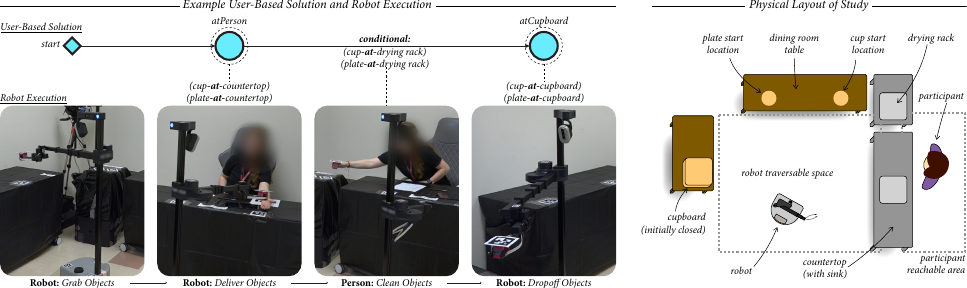}
    \caption{The smallest solution to the \textit{tidying} scenario and its execution (left). The physical layout of the study room (right).}
    \label{fig:study_setup}
\end{figure*}

To evaluate our systems-level contribution and understand the interaction between the core components of the system, we conducted an IRB-approved laboratory study that compares the full version of \tool{} with an ablated baseline without the \planvisualizer{}.
Our hypotheses are that exposure to the \planvisualizer{} improves the quality of the resulting plans (H1), helps match user expectations to robot task performance (H2), improves the perceived competence of the robot (H3), and improves \tool{}' usability (H4).

\subsection{Study Design}

We conducted an experiment with two conditions---\textit{plan-vis}, in which participants were exposed and allowed access to \tool{}' \planvisualizer{}, and \textit{no-vis}, in which participants were neither exposed nor allowed access to the \planvisualizer{}.
Participants specified a goal automaton and executed the resulting plan on a robot.

\subsubsection{Study Scenario}

Our evaluation centered on a \textit{tidying} scenario.
Participants were informed the following:

\begin{quotation}
  \textit{You are finished hosting a dinner for some friends, and now it is time to clean up. While you wash dishes in the kitchen, you want your robot to help deliver dirty dishes to you and deliver clean dishes to the cupboard.}
\end{quotation}

Figure \ref{fig:study_setup} (right) depicts the physical layout of the study.
Participants were informed that their job was to wash a dirty plate and cup.
Participants were also informed of the robot's capabilities---it can deliver dishes to and from the participant's vicinity and open the cupboard.
Participants were not allowed to step out from behind the countertop (see \textit{participant reachable area} in Figure \ref{fig:study_setup}, right).

Participants were informed that to clean a dish, they needed to access the dish (presumably by having the robot deliver the dish to their vicinity) and place the dish on the drying rack.
Once on the drying rack, the dish is clean and ready to be put away.
The smallest solution for the \textit{tidying} scenario is depicted in Figure \ref{fig:study_setup} (left).
On execution of this solution, the robot delivers the dishes to the countertop, waits until the dishes are on the drying rack, then opens the cupboard, and finally moves the dishes to the cupboard.

\subsubsection{Measures}

To measure plan quality, we first enumerate four basic objectives of the \textit{tidying} scenario---(1) cup clean, (2) plate clean, (3) clean cup in the cupboard, and (4) clean plate in the cupboard.
Giving each objective equal weight, we then compute (1) a \textit{runtime score}, or how many objectives the robot meets during plan execution; and (2) a \textit{feasibility score}, or the maximum number of objectives that the robot \textit{could} meet at runtime. The \textit{runtime} score may be lower than the \textit{feasibility} score if the participant acts suboptimally at runtime, \eg{} if the participant removes the cup and plate from the drying rack before the robot has had the chance to grab them. Higher values are better for \textit{runtime} and \textit{feasibility} scores. We additionally created (3) a third and more fine-grained analysis of task quality---\textit{human effort}. Given a participant's task plan, the measure asks: What is the minimum number of independent actions that a human would have to perform during the robot's execution for all four objectives in the task to be achieved? To compute this measure, we relax the assumption that participants stay behind the countertop. Lower values of \textit{human effort} are better.

We measure usability via the SUS questionnaire (10 items, 5-point Likert scale) \cite{brooke1996sus} and the usefulness (8 items), ease of learning (4 items), and satisfaction (7 items) factors of the USE questionnaire (7-point Likert scale) \cite{lund2001measuring}. To measure perceived robot competence, we include the competence factor of the RoSAS scale (6 items, 7-point Likert scale) \cite{carpinella2017robotic}.
We developed our own \textit{expectations} questionnaire to measure the degree to which expectations of robot performance are matched in terms of four factors (5 items each, 7-point Likert scale)---expectations \textit{overall} (Cronbach's $\alpha=0.80$), expectations of \textit{what} the robot did (Cronbach's $\alpha=$0.78), and expectations of \textit{where} (Cronbach's $\alpha=0.91$) and \textit{why} (Cronbach's $\alpha=0.87$) it did it.
Although not part of our hypotheses, we measured task load through the NASA TLX (7 items, 7-point Likert scale) \cite{hart1988development}.\footnote{Copies of the study materials can be found at \url{https://osf.io/ewfd5/}.}

\subsubsection{Procedure}

Study sessions lasted for one hour.
After giving their consent to participate, participants completed a self-guided browser-based \tool{} tutorial.
Participants were encouraged to ask questions at this stage, to which the experimenter responded within the scope of the tutorial.
\textit{Plan-vis} participants were exposed to the \planvisualizer{} through an additional tutorial step.

After the tutorial, participants were briefed on the \textit{tidying} scenario and given 10 minutes to specify a goal automaton within \tool{}.
Within the 10 minutes, \textit{plan-vis} participants had unlimited access to the \planvisualizer{} at their discretion. 
After 10 minutes or when participants indicated that they had finished, we administered the TLX, USE, and SUS questionnaires.

Participants were then given instructions for executing their plans on the robot.
Figure \ref{fig:study_setup} (left) depicts a sample execution.
During execution, participants observed the status of the robot on the tablet. 
If the robot encountered a conditional, it waited for confirmation from the participant before proceeding. 
Participants were informed that they could move any items around in their vicinity, \eg{} to or from the countertop and the drying rack.

During execution, deviations from the robot's expected world state were engineered to cause the robot to prematurely halt execution.
Deviations could occur, for example, if the participant failed to specify in their goal automaton that clean dishes would be placed on the drying rack (\ie{} by failing to insert a conditional that instructs the robot to wait until the dishes are on the drying rack before proceeding to put them in the cupboard).
In this case, the robot would believe the dish to be on the countertop, but without seeing the dish on the countertop, it would be unable to proceed.
Participants would be given a few seconds to realize their mistake and put the clean dish back on the countertop, but if deviations remained uncorrected, the robot halted execution.

At the end of execution, participants filled out the questionnaires for expectations matched and perceived competence.
To adhere to the one-hour time limit, one participant was administered these questionnaires prior to when the robot had finished execution. 
If time permitted, participants underwent semi-structured interviews.
Interview durations varied based on the remaining time in the hour.

\subsection{Results}

\begin{figure}[t!]
    \includegraphics[width=\columnwidth]{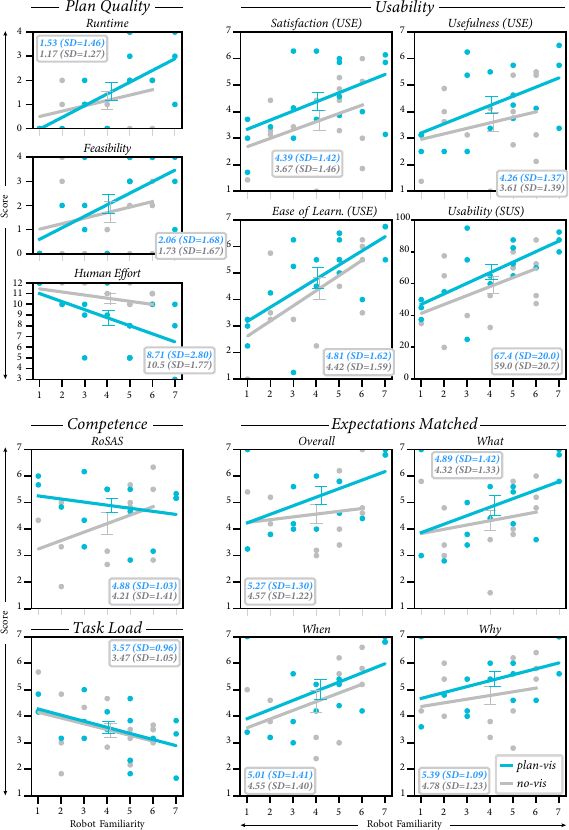}
    \caption{Plan quality (top left), usability (top right), competence and task load (bottom left), and expectations matched (bottom right) plotted against robot familiarity. Linear trendlines indicate the direction of the relationship. Blue is \planvis{}. Grey is \novis{}. Lower is better only for \textit{human effort} and task load. Error bars represent standard error.}
    \label{fig:scatter}
\end{figure}

\paragraph{Participants. }
We recruited 33 volunteers (19 male, 14 female) from within the U.S. Naval Research Laboratory in Washington, D.C. 
We discarded one participant's data (\textit{plan-vis}) due to a rare but experience-altering software bug.
Of the remaining 32 participants (17 \planvis{}, 15 \novis{}), the average age was 30.7 years ($SD=11.7$) and the average self-reported familiarity with robots was 4.09 ($SD=1.89$) on a seven-point, single-item Likert scale (low=1, high=7). 
Six participants reported participating in past robotics studies.

Within our sample, 14 participants (8 \planvis{}, 6 \novis{}) produced plans that were either correct or nearly correct (\textit{feasibility} score of 3-4).
Two participants (1 \planvis{}, 1 \novis{}) created goal automata that failed to compile.
An additional three participants (1 \planvis{}, 2 \novis{}) experienced equipment failure at the time of plan execution.  
Due to these five participants not executing their plans on the robot, our post-execution measures (competence, expectations matched, and \textit{runtime} score) include 27 participants in total.

\paragraph{Hypothesis Testing}
We performed one-tailed Mann-Whitney U tests to compare plan quality and one-tailed Student's t-tests to compare usability, competence, and expectations matched between conditions.
We observed a significant difference in plan quality in terms of \textit{human effort} between the \planvis{} and \novis{}     conditions ($U=182.5$, $p=0.017$). We observed marginal effects ($p<0.1$) for \planvis{} performing better than \novis{}     participants for our measures of perceived usefulness of \tool{} ($p=0.097$, $t(30)=1.33$), satisfaction with \tool{}     ($p=0.081$, $t(30)=1.43$), perceived competence of the robot ($p=0.083$, $t(25)=1.43$), and expectations matched both     \textit{overall} ($p=0.082$, $t(25)=1.43$) and for \textit{why} the robot acted ($p=0.095$, $t(25)=1.35$).     Additionally, \textit{plan-vis} participants reported SUS scores of 67.35 ($SD=19.99$), whereas \textit{no-vis}     participants reported SUS scores of 59.00 ($SD=20.68$).
In the other measures that we compared, the \planvis{} condition generally performed better on average than the \novis{} condition within our sample (Figure \ref{fig:scatter}).
Average values for each measure under both conditions can be found in Figure \ref{fig:scatter}.
These results support the \planvisualizer{} in increasing overall plan quality, but further investigation is required before accepting our hypotheses.

\paragraph{Robot Familiarity}
We additionally analyzed each of our measures for associations with self-reported robot familiarity.
Table \ref{table:corr} (left) shows the resulting Spearman's rank correlations, and Figure \ref{fig:scatter} visualizes these associations for each condition.
It can be seen that as robot familiarity increases, plan quality, usability, and expectations matched also increase, while task load decreases.
These correlations suggest that end-user programmers' \color{black} past familiarity with robots may impact almost every interaction that they have with \tool{}.
Our manipulation may be competing with robot familiarity.

\begin{table}
\small
\centering
\begin{tabular}{|r|cc|cc|}
 \hline
 & \multicolumn{2}{c|}{Correlation with} & \multicolumn{2}{c|}{Correlation with} \\
 & \multicolumn{2}{c|}{Robot Familiarity} & \multicolumn{2}{c|}{\planvisualizer{} Usage} \\
 \hline
 \textbf{Measure} & \textbf{Spearman's $r$} & \textbf{$p$} & \textbf{Spearman's $r$} & \textbf{$p$} \\
 \hline

    \textit{Runtime} score & \textbf{0.556}\color{black} & \textbf{<0.01} & 0.468\color{black} & 0.079\color{black} \\

    \textit{Feasibility} score & \textbf{0.445}\color{black} & \textbf{0.011}\color{black} & 0.387\color{black} & 0.125\color{black} \\

    \textit{Hum. effort} cost & \textbf{-0.460}\color{black} & \textbf{<0.01}\color{black} & -0.377\color{black} & 0.136\color{black} \\

    Expect. \textit{overall} & \textbf{0.398} & \textbf{0.040} & \textbf{0.831}\color{black} & \textbf{<0.001} \\

    Expect. \textit{what} & \textbf{0.422} & \textbf{0.028} & \textbf{0.737}\color{black} & \textbf{<0.01} \\

    Expect. \textit{when} & \textbf{0.484} & \textbf{0.011} & \textbf{0.784}\color{black} & \textbf{<0.001} \\

    Expect. \textit{why} & 0.371 & 0.057 & \textbf{0.614}\color{black} & \textbf{0.015}\color{black} \\

    Competence & 0.066\color{black} & 0.745\color{black} & 0.051\color{black} & 0.857\color{black} \\

    Usability (SUS) & \textbf{0.568}\color{black} & \textbf{<0.001}\color{black} & \textbf{0.496}\color{black} & \textbf{0.043}\color{black} \\

    Usefulness & \textbf{0.398}\color{black} & \textbf{0.024}\color{black} & 0.433\color{black} & 0.083\color{black} \\

    Ease of Learn. & \textbf{0.639} & \textbf{<0.001}\color{black} & 0.453\color{black} & 0.068\color{black} \\

    Satisfaction & \textbf{0.390} & \textbf{0.027} & 0.226\color{black} & 0.383\color{black} \\

    Task Load & \textbf{-0.399} & \textbf{0.024}\color{black} & -0.092\color{black} & 0.724\color{black} \\
 \hline
\end{tabular}
\caption{Spearman's rank coefficient with robot familiarity (all data) and \planvisualizer{} usage (\planvis{} only) for each measure. Bold indicates statistical significance ($p<0.05$).}
\label{table:corr}
\end{table}

\paragraph{Plan Visualizer Usage}
We grouped \textit{plan-vis} participants (referred to as P\textit{X}, with \textit{X} being a unique identifier) into various categories based on how they used the \planvisualizer{}.
We categorized eight participants (P2, P5, P6, P7, P22, P25, P27, and P29) as ``\textit{intended use}.'' These participants appeared to use the \planvisualizer{} to validate their work or guide them in fixing errors and produced plans of high quality (feasibility score mean of $3.625$ out of $4$). 
We categorized an additional two participants (P16 and P30) as ``\textit{unsuccessful use}'' due to heavy reliance on the \planvisualizer{} but an inability to fix errors in their goal automata (feasibility score mean of $0.5$ out of $4$). 
Four more participants (P4, P9, P13, and P21) fall into the ``\textit{no-use}'' category due to not accessing the \planvisualizer{} at all or accessing it but not interacting with it (\eg{} by not scrolling through or clicking on actions in the plan). \textit{No-use} participants produced plans of low quality (feasibility score mean of $0.75$ out of $4$).
We grouped the remaining three participants into a category of ``\textit{unknown use},'' for whom the role of the \planvisualizer{} is unclear (feasibility score mean of $0.66$ out of $4$).

Within the \planvis{} condition, there are strong positive associations between \planvisualizer{} usage (equal to how many times a participant accessed the \planvisualizer{}, and while accessing it, interacted with the \planvisualizer{} as well) and expectations matched. Table \ref{table:corr} (right) depicts these correlations.
\section{Discussion}
In our experiment, we found evidence that the \planvisualizer{} increases plan quality and marginal effects for the \planvisualizer{} increasing satisfaction, expectations matched, perceived usefulness of \tool{}, and perceived competence of the robot.
On average, the \planvis{} condition generally performed better than the \novis{} condition.
We find these results encouraging, but further investigation is required to fully understand the \planvisualizer{}'s effectiveness, and more generally, \tool{} overall.
To guide our further investigation and provide guidance to future research within the wider EUP community, we propose three design implications. 

\paragraph{\textbf{Design Implication:} Feedback is critical for goal-oriented EUP} This implication is evidenced by the significant and marginal effects of the \planvisualizer{} despite its underuse by \textit{``no use''} and \textit{``unknown use''} users.
There also exists a strong positive association between \planvisualizer{} usage and both expectations matched and usability.
Although this association is not causal, we believe that improving \planvisualizer{} access would have increased our observed effect.
The importance of feedback and how information is presented to users is further supported by prior work \cite{porfirio2018authoring, amershi2019guidelines}.
\textit{\textbf{Recommendation:} Feedback should be provided proactively (rather than passively) by goal-oriented EUP tools.} \tool{} users should be exposed to feedback as soon as changes to their programs occur, which could result in higher expectations matched for \textit{``no use''} and \textit{``unknown use''} users.

\paragraph{\textbf{Design Implication:} Although goal-oriented programming allows for greater flexibility in theory (see \S\ref{sec:introduction}) and has shown benefit in prior work \cite{cox2007mixed, hu2012teaching}, users' ability to leverage this flexibility in practice should not be presumed.}
As evidence of this implication, less than half of the participants in either condition produced correct or nearly correct plans.
Participant interviews reveal a potential explanation: Goal predicates are difficult to reason about and require a shift in thinking from a potentially more intuitive (albeit less flexible) action-oriented paradigm (P9, P17, P19).
\textit{\textbf{Recommendation:} EUP researchers must investigate user-interface techniques that improve user comprehension of goal-oriented programming.} Crucially, EUP researchers should avoid assuming that current interface norms for EUP seamlessly translate to the goal-oriented paradigm.

\paragraph{\textbf{Design Implication:} Robot familiarity strongly predicts perceptions and use of robot EUP tools. }
As evidence of this implication, our evaluation uncovers significant correlations between robot familiarity and usability, expectations matched, plan quality, and task load. 
We note that our study population is critical to uncovering this finding---our sample includes both professionals and students at both ends of the spectrum of robot familiarity.
At the same time, the observed effects of our manipulation are potentially diluted due to the breadth of our study population. 
\textit{\textbf{Recommendation:} EUP researchers for human-robot interaction need to choose their study population more carefully and deliberately than is often done in current practice.} Robot familiarity should factor into this choice.

\paragraph{Limitations and Future Work}

\tool{}' design poses various opportunities for improvement.
Most notably, users need assistance creating ``correct'' plans (\ie{} plans with high feasibility scores).
Future work should thus employ formal methods to help improve plan feasibility, such as by automatically detecting and fixing contradictory goals.
Other limitations include that \color{black} our planning approach does not account for uncertainty, such as if the location of an entity in the task context (\eg{} a person) is unknown or if there is an unknown number of multiple items of the same type.
\tool{}' inability to support loops or enumeration (\eg{} tasking the robot to deliver food to \textit{all} rooms in the care facility) further limits the task contexts within which it can operate.
For greater applicability in the wild, \tool{} can also support plan adaptation, \eg{} by learning action costs and re-planning at runtime.

Further limitations exist in our evaluation.
Primarily, our study is systems-level, focusing on the interaction between core components rather than exploring the benefit of increased flexibility from goal-oriented programming.
Future component-level testing is already in preparation to understand the benefit of flexibility in practice.
Additionally, we tested \tool{} with just one scenario and a broad user group and did not collect data about how much training time is required for \tool{}.
Although our sample is critical to revealing significant associations with robot familiarity, future work must explore \tool{} with its target user base, more realistic scenarios, and explore approaches to user training.
We believe that our present study provides an excellent foundation for future testing, such as by deploying \tool{} \textit{in situ} with actual caregivers.

\section{Conclusion}
We present \tool{}, a novel goal-oriented end-user programming (EUP) system.
The purpose of \tool{} is to provide flexibility to robot end users in the level of detail that programs are specified while ensuring that user expectations match robot performance.
Our evaluation of \tool{} uncovers evidence that plan feedback increases the quality of user-created programs. The evaluation also uncovers strong associations between plan feedback, robot familiarity, and participant experience and performance.
We conclude with various design implications for the future development of EUP tools.

\begin{acks}
This research was supported by the Office of Naval Research and an NRC Research Associateship award to DP at the U.S. Naval Research Laboratory. The views and conclusions contained herein are those of the authors and should not be interpreted as necessarily representing the official policies, either expressed or implied, of the U.S. Navy. We thank Greg Trafton for his input on our evaluation.
\end{acks}

\bibliographystyle{ACM-Reference-Format}
\balance
\bibliography{biblio}

\end{document}